# Interpretable Hierarchical Attention Network for Medical Condition Identification

Dongping Fang
*AI Data Science*
*Elevance Health, Inc.*
Indianapolis, USA
dongping.fang@carelon.com

Lian Duan
*AI Data Science*
*Elevance Health, Inc.*
Indianapolis, USA
lian.duan@carelon.com

Xiaojing Yuan
*AI Data Science*
*Elevance Health, Inc.*
Indianapolis, USA
xiaojing.yuan@carelon.com

Allyn Klunder
*AI Data Science*
*Elevance Health, Inc.*
Indianapolis, USA
allyn.klunder@carelon.com

Kevin Tan
*AI Data Science*
*Elevance Health, Inc.*
Indianapolis, USA
kevin.tan@carelon.com

Suiting Cao
*AI Data Science*
*Elevance Health, Inc.*
Indianapolis, USA
suiting.cao@carelon.com

Yeqing Ji
*AI Data Science*
*Elevance Health, Inc.*
Indianapolis, USA
yeqing.ji@carelon.com

Mike Xu
*AI Data Science*
*Elevance Health, Inc.*
Indianapolis, USA
xiaoming.xu@carelon.com

*Abstract*—Accurate prediction of medical conditions with straight past clinical evidence is a long-sought topic in the medical management and health insurance field. Although great progress has been made with machine learning algorithms, the medical community is still skeptical about the model accuracy and interpretability. This paper presents an innovative hierarchical attention deep learning model to achieve better prediction and clear interpretability that can be easily understood by medical professionals.

This paper developed an Interpretable Hierarchical Attention Network (IHAN). IHAN uses a hierarchical attention structure that matches naturally with the medical history data structure and reflects patients' encounter (date of service) sequence. The model attention structure consists of 3 levels: (1) attention on the medical code types (diagnosis codes, procedure codes, lab test results, and prescription drugs), (2) attention on the sequential medical encounters within a type, (3) attention on the individual medical codes within an encounter and type.

This model is applied to predict the occurrence of stage 3 chronic kidney disease (CKD), using three years' medical history of Medicare Advantage (MA) members from an American nationwide health insurance company. The model takes members' medical events, both claims and Electronic Medical Records (EMR) data, as input, makes a prediction of stage 3 CKD and calculates contribution from individual events to the predicted outcome.

## I. Introduction

Predictions of patients' risks of having certain medical conditions are very useful for early diagnosing, preventing, monitoring, treating, and budgeting in the healthcare industry. Predictive analytics can turn raw healthcare data into practical insights for enhancing health plan decision-support and education tools for providers, provider's patient care and overall operational effectiveness. The commonly used prediction algorithms are machine learning algorithms like logistic regression, random forest, lightGBM, etc. that take inputs in table format with rows representing patients, and columns representing features of the patients. A patient's historical medical records are timely ordered medical codes. It usually takes a lot of effort to engineer features from these medical codes. In addition, the engineered features cannot possibly represent all information captured by these medical codes. Also, heavily engineered features are hard to understand and interpret.

IHAN is very flexible in its inputs: timely ordered sequential data for example. It's particularly fit for patients' medical records. A patient may visit a doctor on any day, for any medical conditions, may have different medicines and lab tests. Lack of interpretability is one of the criticisms of deep learning models. Interpretability helps people to understand the reasons and gains insights into how things work. Interpretability is even more important in the medical field. Developing a new interpretable deep learning model directly working on all medical codes is the topic of this paper.

An important work that we drew inspirations from is the RETAIN paper by Choi et al [1]. They directly address the interpretability of deep learning predictive models. In their paper, two layers of attentions are used: one is on visit (medical encounter), and the other one is on coordinate of visit embedding, and none is used directly on individual codes within a visit. They also only used EMR data from one medical group.

From our investigation of a major source for EMR data, we find that EMR's data quality and completeness are limited. On the other hand, claims data are much more reliable and complete. The medical codes considered in RETAIN are, diagnosis, medication, and procedure codes; lab code was not used. In addition, the medical codes used in RETAIN are not at the raw level, instead, they are medical groupers with multiple codes aggregated into input variables. We believe that code at its raw level can provide interpretation that's more detailed and easier to understand.

Other relevant works are in [1-8] where at most 3 types of medical codes, and at most two layers of attentions are used. In the published papers, majority used EMR or EHR from a single hospital/group/foundation since the claims data is not available to most researchers. The EMR data used the most is the MIMIC data. MIMIC data are for intensive care and hence only represent a small portion of medical data and population.

This paper uses all data described in the next section from a nationwide health insurance company. Health insurance company is in a unique position to have the most complete data due to, a) claims data is owned by health insurance company that is not usually available to academic researchers; b) health insurance company acquires large amount lab test data and EMR data of their members across all sources not just a few providers, groups, or facilities. This makes this paper one of the studies using much more data than other studies did.

## II. MEDICAL DATA

### A. Claim Data

Claim data includes the information reported by healthcare providers (including physicians, hospitals and pharmacies) to health insurance companies for payment purposes. Claims data contains different types of medical codes, describing the health services offered to the patients, including diagnosis codes, procedure codes, drug codes, etc. In addition, claims data also provide the information on date of service, servicing provider specialties, place of service, patient demographics, as well as charges of the services.

### B. Lab Data

Lab data are maintained by laboratory facilities to document the details of patients' laboratory test encounters, including test order dates, lab test codes or panel codes, lab results, etc.

### C. EMR

Electronic Medical Records (EMR) or Electronic Health Records (EHR) are confidential records that are created by healthcare providers to document patient care. patient demographics, medical history, and documentation of symptoms, diagnosis, treatment, and outcome are contained in the EMR systems of healthcare providers.

## III. ALGORITHM

We will use medical data to set up our notations and terms. For a patient or member (patient and member are used interchangeably in the paper), let $\{x_{ijk}: i = 1, \ldots, I; j = 1, \ldots, J_i; k = 1, \ldots, K_{ij}\}$ denote the $k$-th code of the $j$-th encounter (or visit) for the $i$-th code type. There are 4 code types considered here: diagnosis, procedure, lab test, and medication. Encounter is referred to as a collection of medical events that happened on the same date of service. For example, an office visit to a doctor results in a collection of diagnoses and medication codes. Patients do not have encounters every day. The number of encounters changes with patient. Within each encounter, there may be multiple medical codes of one or more code types. The number of codes and code types change for each encounter. The encounters are timely ordered but at irregular time intervals. Within one encounter, the order of the codes documented doesn't matter to our model.

Our algorithm uses a hierarchical attention structure that matches naturally with the medical history data structure and reflects the patient's encounter (date of service) sequence. The model attention structure consists of 3 levels: (1) attention on the medical code types (diagnosis codes, procedure codes, lab test results, and prescription drugs), (2) attention on the sequential medical encounters within a type, (3) attention on the medical codes within an encounter and type.

We present the algorithm for one type of code first, then the algorithm for multiple types of codes.

### A. Algorithm for single type of medical code

Fig. 1 illustrates our model for a single type of medical code, the diagnosis code for example. Since there is only one code type, I =1, index $i$ is dropped: $\{x_{jk}: j = 1, \ldots, J; k = 1, \ldots, K_j\}$. Following steps are performed.

1. The input of medical codes for each encounter
    encounter 1: $x_{11}, x_{12}, \ldots, x_{1K_1}$
    encounter 2: $x_{21}, x_{22}, \ldots, x_{2K_2}$
    …

2. Medical code embedding
    $$\mathbf{e}_{jk} = W^{(emb)} f(x_{jk})$$
    In this paper, each medical code is embedded as a vector of size 128. Embedding matrix $W^{(emb)}$ is a parameter to be learned. The $f(.)$ is the one-hot-encoding function that transforms input to a binary vector with all 0s except one 1 at the location of corresponding medical code.

3. Medical code attentions within $j$-th encounter
    $$\alpha_{j1}^{(c)}, \ldots, \alpha_{jK_j}^{(c)} = \text{softmax}(W_c \mathbf{E}_j + b_c)$$
    where $\mathbf{E}_j = (\mathbf{e}_{j1}, \ldots, \mathbf{e}_{jK_j})$

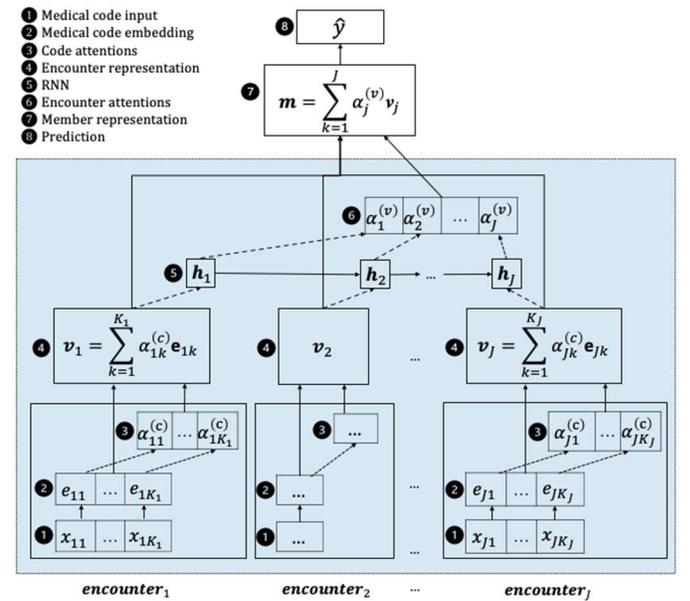

Fig. 1. Architecture for single medical code type, diagnosis code for example. Two level attention layers are used: the code attention layer and the visit attention layer.

4. Encounter (or visit) representation
$$v_j = \sum_{k=1}^{K_j} \alpha_{jk}^{(c)} e_{jk}$$

5. Sequential model GRU
$$H = (h_1, h_2, ..., h_J) = GRU(v_1, v_2, ..., v_J)$$
In this paper, we used model Gated Recurrent Unit (GRU), but any proper model like LSTM, transformer, etc., will work as well.

6. Encounter (or visit) attentions
$$\alpha_1^{(v)}, ..., \alpha_J^{(v)} = \text{softmax}(W_v H + b_v)$$

7. Member representation
$$m = \sum_{j=1}^{J} \alpha_j^{(v)} v_j$$

8. Prediction
$$\hat{y} = \text{sigmoid}(Wm + b)$$

### B. Algorithm for multiple types of medical codes

We present two approaches to combine different types of medical codes.

*1) Algorithm level combination of different code types*

This approach applies single code algorithm on each code type to get member representation of each code type first, and then use another attention layer to combine different code types together (Fig. 2).

Detailed steps are:

1. Member representation for each code type, using the single code type model step 1 to 7 to calculate:
$$m_1, m_2, ..., m_I$$

2. Medical code type attentions
$$\alpha_1^{(t)}, ..., \alpha_I^{(t)} = \text{softmax}(W_t M + b_t)$$
Where $M = (m_1, m_2, ..., m_I)$

3. Member representation
$$m = \sum_{i=1}^{I} \alpha_i^{(t)} m_i$$

4. Prediction
$$\hat{y} = \text{sigmoid}(Wm + b)$$

*2) Data level combination of different code types*

This approach combines all different types of medical codes at the input data level, i.e., simply concatenate together all different types of codes that occurred on the same date of service (encounter). Then this reduces to the single code type situation. Doing this puts codes of different types in the same position. There is no concept of code type anymore.

Data level combination (data_comb) is a common approach used by other researchers when using multiple medical code types [1-5, 7]. The data_comb approach uses more memory during model training due to bigger set of combined codes compared to algorithm level combination (algorithm_comb). We have not found other researchers using algorithm_comb approach described in section III.B.1. The algorithm_comb uses less memory during training and is more intuitive to summarize the contributions of different code type. Due to its componentized approach, algorithm_comb is easier to take advantage of distributed computation.

### C. Contribution coefficient for interpretation

When we give a prediction of a patient's risk of having a certain medical condition, we would also like to know why. We derive the contribution of each medical code to the prediction by going back applying the equations in above steps,

$$\hat{y} = \text{sigmoid}(Wm + b) = \cdots$$
$$= \text{sigmoid}\left(\sum_{i=1}^{I} \sum_{j=1}^{J_i} \sum_{k=1}^{K_{ij}} w_{ijk} + b\right)$$

Where $w_{ijk}$ is the contribution coefficient from the $i$-th type, $j$-th encounter, $k$-th code,

$$w_{ijk} = \alpha_i^{(t)} \alpha_{ij}^{(v)} \alpha_{ijk}^{(c)} W W_i^{(emb)} f(x_{ijk}).$$

For single code type, contribution coefficient from the $j$-th encounter, $k$-th code,

$$w_{jk} = \alpha_j^{(v)} \alpha_{jk}^{(c)} W W^{(emb)} f(x_{jk})$$

### D. Parameters to be trained

For the model of single code type, the parameters are: $W^{(emb)}$, $W_c, b_c, W_v, b_v$, parameters in GRU, $W, b$. For the model of multiple code types, there are multiple sets of ($W^{(emb)}$, $W_c, b_c, W_v, b_v$, parameters in GRU) one per code type, $W_t, b_t$, and $W, b$.

## IV. APPLICATION

### A. Experimental setting

*1) Objective*

Among Medicare Advantage (MA) members from a nationwide health insurance company, we aim to predict the occurrence of stage 3 chronic kidney disease (stage 3 CKD) in 2020, using previous three years' medical history of members from 2017 to 2019.

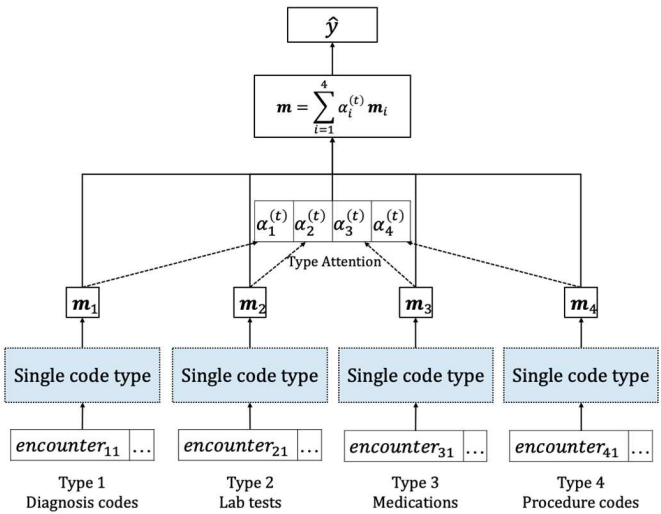

Fig. 2. Architecture for multiple types of medical codes, diagnosis, medication, lab test, and procedure code for example. The blue boxes represent the single code type architecture (blue box) in Fig. 1. Type attention layer is added to consolidate the effects from multiple types.

### 2) Data

The historical medical claims, pharmacy claims, and lab data of the MA members were obtained from the company's Enterprise Data Warehouse. Medical claims were used to extract members' date of service, diagnosis codes (ICD10-CM [10]), outpatient procedure codes (CPT4/HCPCS [11]), and inpatient procedure codes (ICD10-PCS [12]). Pharmacy claims were used to extract the filled dates and GPI [13] codes of members' drug prescriptions (GPI8 codes, first 8 digits of GPI, were used). Laboratory test dates, laboratory codes (LOINC [14]) and test abnormality codes were obtained from the lab data. For lab data, we further combined LOINIC code and abnormality codes as our input lab test code, for example, '88294-4_L' for LOINIC ='88294-4' and abnormality code = 'L'. The combined lab code '88294-4_L' means test 'glomerular filtration rate in the blood' is performed and the result is in the abnormally low range.

The company has access to the EMR extracts of a substantial number of its network hospital systems and physician organizations. In addition to the above claim and lab data, more data on patient's historical date of service, diagnosis codes (ICD10-CM), procedure codes (CPT/HCPCS), prescription drug codes (GPI8), lab codes (LOINC), and test abnormality codes were also extracted from the EMR for available patients.

The inputs for IHAN were prepared to the specific date of service, code type and individual code level for each MA member.

For comparison, we fitted some machine learning models based on the same information. Since the usual machine learning model takes inputs from a tabular format table, we engineered more than 10,000 features based on the four types of medical codes. These features were prepared at the yearly level and aggregated to certain grouper of each code type. ICD10-CM diagnosis codes were grouped using the Clinical Classification Software Refined (CCSR, 528 categories) developed by Healthcare Cost and Utilization Project (HCUP), CPT/HCPCS procedure codes were grouped into procedure groups maintained by the company's Enterprise Data Analytics team (1,815 categories for CPT/HCPCS code), ICD10PCS procedure codes were grouped into the Procedure Classes Refined for ICD-10-PCS (547 categories), and GPI drug codes were aggregated at GPI4 level (1,510 categories). Number of claims from each code category was calculated on yearly basis. The legacy model also used the count of abnormal HbA1C in each year as input features.

### 1) Details

In 2020, 985,413 members were continuously enrolled in MA plans of a nationwide health insurance company. For claims data, members who had at least 2 encounters with any claimed diagnosis codes in the 3 years from 01/01/2017 to 12/31/2019 were included in the model population (N=800,607). Among them, 9.15% or 73,242 members had label = 1 (case). To balance the data, we kept all members with label = 1, and randomly selected members with label = 0 (non-case), making the case to the non-case ratio of 1:3. This reduced the member count to 292,968. Similarly, we applied the same filter and label balance on EMR data. For combined Claim+EMR data, we added available EMR data to the claims data for the patient population from the claims data. Table I is the summary of data used in our model development.

The data was further split into training (60%), validation (20%) and testing (20%) datasets. The models were trained using the training dataset and the validation datasets were used to stop training when the loss on the validation data did not decrease any more. Finally, the trained models were applied on the testing dataset to provide a fair estimate of model performance.

Table I shows that the claim data is much more complete than EMR data except for the lab test codes. For lab test, EMR brings a lot of extra codes.

To train models, we used PyTorch's BCELoss (Binary Cross Entropy) loss function, and AdamW optimizer with initial learning rate = 5e-4. The dimension of embedding of medical code is 128. All our computations were done using one NVIDIA T4 GPU.

The patient population is much bigger, the records and types of medical data are more comprehensive in our work.

### B. Results

Taking the advantage of rich and valuable medical data from a nationwide health insurance company, we performed various comparisons to show the benefit of including different data sources (EMR, Claim) and different types of medical codes (Diag, Lab, Rx, Proc) in terms of predicting stage 3 CKD. These comparisons can show us how much we might have missed if we ignored some data source or type. We see other researchers used at most 3 types of medical codes, and lab test results were not commonly used in model development.

TABLE I. MODEL DATA SUMMARY

|  | Claim | EMR | Claim+EMR |
|---|---|---|---|
| # members | 292,968 | 160,772 | 292,968 |
| # encounters per member, median | 57 | 10 | 59 |
| # codes per encounter, median | 3 | 3 | 3 |
| # unique diag codes | 42,683 | 27,921 | 45,937 |
| # unique proc codes | 29,231 | 8,008 | 30,082 |
| # unique lab test codes | 12,263 | 23,402 | 27,981 |
| # unique rx codes | 1,879 | 1,726 | 2,194 |

TABLE II. MODEL PERFORMANCE: AVERAGE TEST AUC AND STANDARD DEVIATION (OVER 20 RUNS)

|  | Data Source | diag | diag+lab | diag+lab+rx | diag+lab+rx+proc |
|---|---|---|---|---|---|
| IHAN | EMR | 0.7866 ±0.0033 | 0.8251 ±0.0105 | 0.8249 ±0.0095 | 0.8275 ±0.0029 |
|  | Claim | 0.8765 ±0.0049 | 0.8981 ±0.0035 | 0.8979 ±0.0051 | 0.8999 ±0.0044 |
|  | Claim+EMR | 0.8774 ±0.0053 | 0.9017 ±0.0054 | 0.9058 ±0.0036 | 0.9071 ±0.0033 |
| Logistic Regression | Claim |  |  |  | 0.8141 ±0.0003 |
| LightGBM | Claim |  |  |  | 0.8892 ±0.0002 |

*1) Model performance*

Diagnosis codes are the most available and most important type. We used diagnosis codes alone to fit a model as our reference model. We then added lab test results, medication codes and procedure codes stepwise, in the listed order, to evaluate the importance of each added code type. We added the procedure codes last because we believe that with given diagnosis codes, procedure codes may not add much extra values compared to lab and medications.

We used test data area under ROC curve (AUC) to measure model performance. In practice, it is hard to draw conclusions on which is better when the difference is small. Many factors may influence test data AUC, for example, the split of data into train/validate/test datasets, the data order we feed to algorithm when training, the learning rate, the initial parameter values, etc. To make the comparisons more reliable, we fitted the same model 10 times for each combination of data source (EMR, Claim, Claim + EMR), code types (diag, diag + lab, diag + lab + rx, diag + lab + rx + proc) and code type combination approach (algorithm_comb, data_comb). We then calculated the average AUC and its standard deviation, and performed statistical two-sample t-tests to test the difference in model performance.

For comparison, we also built logistic regression models and lightGBM models using 10,000+ features (described in experiment setting section) engineered based on the four types of medical codes.

Table II shows average test AUC for model built with different data sources and code types. Table III to VI show the two-sample t-test for evaluating the effects of code type, data sources, and two approaches combining different code types.

We draw following conclusions from Table II to VI. When t-test is performed, we tested at significance level of 0.05. The difference is considered statistically significant if p-value < 0.05.

1. IHAN vs traditional machine learning models (Table II – III).

    1.1. The AUC from logistic regression is much lower than the equivalent IHAN models.

    1.2. AUC from lightGBM is good but still about 1% lower than IHAN.

2. Effects of different code types (Table IV)

    2.1. lab test – adding lab test results significantly increased the model performance, with a 2% to 4% increase in AUC across data sources.

TABLE III. TWO-SAMPLE T-TEST FOR COMPARING IHAN MODEL AND TRADITIONAL MODELS (USE ALL 4 CODE TYPES)

| Data Source | Model 1 | Model 2 | Mean AUC 1 | Mean AUC 2 | p-value |
|---|---|---|---|---|---|
| Claim | Logistic Regression | IHAN | 0.814 | 0.900 | <0.0001 |
| Claim | lightGBM | IHAN | 0.889 | 0.900 | <0.0001 |

TABLE IV. TWO-SAMPLE T-TEST FOR COMPARING IHAN MODELS FITTED WITH DIFFERENT CODE TYPES

| Data Source | Code types 1 | Code types 2 | Mean AUC 1 | Mean AUC 2 | p-value |
|---|---|---|---|---|---|
| EMR | diag | diag+lab | 0.787 | 0.825 | <0.0001 |
| EMR | diag+lab | diag+lab+rx | 0.825 | 0.825 | 0.966 |
| EMR | diag+lab+rx | diag+lab+rx+proc | 0.825 | 0.828 | 0.258 |
| Claim | diag | diag+lab | 0.877 | 0.898 | <0.0001 |
| Claim | diag+lab | diag+lab+rx | 0.898 | 0.898 | 0.892 |
| Claim | diag+lab+rx | diag+lab+rx+proc | 0.898 | 0.900 | 0.197 |
| Claim+EMR | diag | diag+lab | 0.877 | 0.902 | <0.0001 |
| Claim+EMR | diag+lab | diag+lab+rx | 0.902 | 0.906 | 0.006 |
| Claim+EMR | diag+lab+rx | diag+lab+rx+proc | 0.906 | 0.907 | 0.269 |

   2.2. Medication codes (Rx) – the improvement by adding medication codes was insignificant if diagnosis and lab were already included except for the 'Claim + EMR' data source. For the 'Claim + EMR' data source, the addition of Rx increased AUC by about 0.4%, very small but statistically significant.

   2.3. Procedure codes – the incremental improvement in model performance from adding procedure codes to the first three code types was insignificant across different data sources.

3. Values of different data sources (Table V)

   3.1. Models built with claim data alone were significantly better than those built with EMR data alone, by 7% to 9% in AUC, regardless of what code types were used. This is likely due to that claim data is much more complete than EMR data.

TABLE V. TWO-SAMPLE T-TEST FOR COMPARING IHAN MODELS FITTED WITH DIFFERENT DATA SOURCES

| Code types | Data Source 1 | Data Source 2 | Mean AUC 1 | Mean AUC 2 | p-value |
|---|---|---|---|---|---|
| diag | EMR | Claim | 0.787 | 0.877 | <0.0001 |
| diag | Claim | Claim+EMR | 0.877 | 0.877 | 0.568 |
| diag+lab | EMR | Claim | 0.825 | 0.898 | <0.0001 |
| diag+lab | Claim | Claim+EMR | 0.898 | 0.902 | 0.017 |
| diag+lab+rx | EMR | Claim | 0.825 | 0.898 | <0.0001 |
| diag+lab+rx | Claim | Claim+EMR | 0.898 | 0.906 | <0.0001 |
| diag+lab+rx+proc | EMR | Claim | 0.828 | 0.900 | <0.0001 |
| diag+lab+rx+proc | Claim | Claim+EMR | 0.900 | 0.907 | <0.0001 |

TABLE VI. TWO-SAMPLE T-TEST FOR COMPARING IHAN MODELS FITTED USING TWO APPROACHES OF COMBINING MULTIPLE CODE TYPES

| Data Source | Code types | Mean data_comb AUC 1 | Mean algorithm_comb AUC 2 | p-value |
|---|---|---|---|---|
| EMR | diag+lab | 0.828 | 0.822 | 0.256 |
| EMR | diag+lab+rx | 0.828 | 0.822 | 0.169 |
| EMR | diag+lab+rx+proc | 0.829 | 0.826 | 0.118 |
| Claim | diag+lab | 0.899 | 0.897 | 0.094 |
| Claim | diag+lab+rx | 0.898 | 0.898 | 0.971 |
| Claim | diag+lab+rx+proc | 0.900 | 0.900 | 0.803 |
| Claim+EMR | diag+lab | 0.902 | 0.901 | 0.792 |
| Claim+EMR | diag+lab+rx | 0.906 | 0.906 | 0.833 |
| Claim+EMR | diag+lab+rx+proc | 0.906 | 0.908 | 0.310 |

- 3.2. Compared to models based on claim alone, models based on both claims and EMR had slightly higher AUC (by 0.4% to 0.8%, P<0.0001), except for the diagnosis code-alone models. For diagnosis codes alone model, EMR doesn't add significant value if claim is already included. Combined with findings in 2.3, we conclude that EMR's lab test and medication data bring in extra values.
4. Comparison of two approaches of combining multiple code types (Table VI)
    - 4.1. There is no significant difference between algorithm combination and data-level combination.

*1) Interpretation*

We demonstrate the model interpretations at three levels: patient-encounter-code level, patient-code level, and code level. Each level gives different insights.

TABLE VII. PATIENT A (PREDICTED RISK = 0.414): CONTRIBUTION COEFFICIENTS FROM MEDICAL CODES (ONLY ABSOLUTE VALUE > 0.01 ARE SHOWN)

| Date of Service | Code Type | Code Description | Contribution Coeff |
|---|---|---|---|
| 2019-02-18 | lab | High creatinine in the blood | 0.070 ▲ |
| | lab | Low glomerular filtration rate | 0.056 ▲ |
| 2019-03-19 | rx | Amlodipine drug | -0.082 ▼ |
| 2019-03-22 | diag | Age-related nuclear cataract, bilateral | -0.047 ▼ |
| 2019-05-06 | proc | Established patient office or other outpatient visit, 20-29 minutes | -0.437 ▼ |
| 2019-06-10 | lab | High creatinine in blood | 0.099 ▲ |
| | lab | Low glomerular filtration rate | 0.079 ▲ |
| 2019-07-21 | rx | Amlodipine drug | -0.064 ▼ |
| 2019-08-18 | diag | Encounter for general adult medical examination without abnormal findings | -0.396 ▼ |
| | proc | Established patient office or other outpatient visit, 20-29 minutes | -0.133 ▼ |
| 2019-09-15 | lab | High creatinine in blood | 0.159 ▲ |
| | lab | Low glomerular filtration rate | 0.127 ▲ |
| 2019-12-29 | lab | High creatinine in blood | 0.330 ▲ |
| | lab | Low glomerular filtration rate | 0.264 ▲ |

TABLE VIII. CUMULATIVE CONTRIBUTION FOR PATIENT A (ONLY ABSOLUTE VALUE > 0.01 ARE SHOWN)

| Code type | Code Description | Contribution Coeff |
|---|---|---|
| diag | Age-related nuclear cataract, bilateral | -0.047 ▼ |
| | Encounter for general adult medical examination without abnormal findings | -0.396 ▼ |
| lab | High creatinine in blood | 0.657 ▲ |
| | Low glomerular filtration rate | 0.527 ▲ |
| proc | Encounter for general adult medical examination without abnormal findings | -0.570 ▼ |
| rx | Amlodipine drug | -0.147 ▼ |

*a) Patient-Encounter-Code Level Interpretation*

Different patients have different medical history and treatment paths. We use one example patient to demonstrate how the past medical codes contributed to the prediction of stage 3 CKD by the IHAN model. Patient A had 36 medical encounters, during the year 2017 to 2019, including a total of 306 medical codes (133 unique medical codes). The model predicted that patient A had a risk = 0.414 of stage 3 CKD with the following well-contributed clinical events – multiple high 'Creatinine in Serum or Plasma' tests and multiple low kidney functioning 'Glomerular filtration rate' tests. Among the abnormal lab test results, more recent results contributed more to the prediction. The model also indicated regular office visits, with no abnormal findings of medical examinations, and taking proper medications decreased the risk of having stage 3 CKD (Table VII).

*b) Patient-Code Level Interpretation.*

For a patient, the same medical code may occur in multiple encounters at different time. We can sum over encounters to get the cumulative contribution of a code to the prediction (Table VIII). This gives outlined look for a patient.

*c) Code Level Interpretation*

To understand what key medical codes would increase the risk, we aggregated across all MA patients and encounters to find out the mean contribution of each code to the prediction. These mean contributions showed a general picture of what medical codes increase or decrease the risk of having stage 3 CKD.

TABLE IX. TOP 10 RISK-INCREASING MEDICAL CODES

| Code Type | Code Description | Number of Patients | Mean Code Contribution |
|---|---|---|---|
| diag | Chronic kidney disease, stage 3 (moderate) | 104,214 | 1.289 |
| lab | Low glomerular filtration rate | 39,018 | 0.324 |
| lab | High creatinine in blood | 91,821 | 0.287 |
| diag | Chronic kidney disease, stage 2 (mild) | 36,871 | 0.206 |
| diag | Disorder resulting from impaired renal tubular function, unspecified | 382 | 0.144 |
| lab | Low glomerular filtration rate | 47,802 | 0.143 |
| rx | Insulin NPH Isophane & Reg (Human) | 7,038 | 0.141 |
| lab | Low glomerular filtration rate | 33,807 | 0.129 |
| diag | Type 2 diabetes mellitus with other diabetic kidney complication | 15,597 | 0.087 |
| diag | Hypertensive chronic kidney disease with stage 1 through stage 4 chronic kidney disease, or unspecified chronic kidney disease | 85,483 | 0.070 |

Table IX shows the top 10 risk-increasing codes. 'Chronic kidney disease, stage 3 (moderate)' is the target disease we are trying to predict. This just reflected the fact that if a patient had stage 3 CKD in the past, he/she would most likely be diagnosed with stage 3 CKD in future encounters. The model easily predicted this most obvious situation. 'Chronic kidney disease, stage 2 (mild)' may become more severe as the disease progresses and the model reasonably identified it as an important factor. The abnormally low GFR test and abnormally high creatinine in the blood increased the risk. Diabetes-related codes also increased the risk.

## V. Conclusion

We have developed an interpretable hierarchical attention network, IHAN, for raw medical code data and applied it to a real-world data to predict stage 3 CKD. Compared to the commonly used logistic regression and lightGBM, IHAN achieved better performance and provided more meaningful and reasonable interpretations.

By comparing and evaluating the usefulness of different medical code types and data sources, we found that using EMR data alone to fit models had much lower performance than using claim data alone. Therefore in practice, medical personnel should be cautious when using models developed solely by using EMR-sourced features. Adding EMR-sourced lab test results and medication data did improve performance slightly. We found the past lab test results were very important to accurately predicting risk for stage 3 CKD. But lab test results were not commonly used by other researchers in relevant works [1-5, 7-8].

Using IHAN methodology, we can generate detailed history medical codes and events that contributed to the risk of having stage 3 CKD. This information can be used as supportive evidence to help physicians make decisions quickly, which is critical in the model application in real clinical settings.